\documentclass[letterpaper, 10 pt, conference]{ieeeconf}  

\IEEEoverridecommandlockouts                              

\usepackage{graphics} 
\usepackage{amsmath} 
\usepackage{amssymb}  
\usepackage{tikz}
\usepackage[utf8]{inputenc}

\title{\LARGE \bf
Hierarchies of Planning and Reinforcement Learning for Robot Navigation
}

\author{Jan W\"ohlke$^{1, 2}$, Felix Schmitt$^{1}$, and Herke van Hoof$^{2}$
\thanks{$^{1}$Bosch Center for Artificial Intelligence, Renningen, Germany. {\tt\small JanGuenter.Woehlke@de.bosch.com, Felix.Schmitt@de.bosch.com}}%
\thanks{$^{2}$UvA-Bosch Delta Lab, University of Amsterdam, Amsterdam, Netherlands. {\tt\small h.c.vanhoof@uva.nl}}%
}

\usepackage{siunitx}
\usepackage{bm}
\usepackage{mathtools}
\usepackage[center]{caption}
\usepackage{subcaption}
\usepackage{pdfpages}
\usepackage{verbatim}

\usepackage[linesnumbered,ruled,vlined]{algorithm2e}
\SetKwInput{KwInput}{Input}                
\SetKwInput{KwOutput}{Output}              
\DeclareMathOperator{\VI}{VI}
\DeclareMathOperator{\TRPO}{TRPO}
\DeclareMathOperator{\epsilongreedy}{epsilon-greedy}
\DeclareMathOperator{\Adam}{Adam}

\newcommand\copyrighttext{%
	\footnotesize $\copyright$ 2021 IEEE.  Personal use of this material is permitted.  Permission from IEEE must be obtained for all other uses, in any current or future media, including reprinting/republishing this material for advertising or promotional purposes, creating new collective works, for resale or redistribution to servers or lists, or reuse of any copyrighted component of this work in other works. DOI: 10.1109/ICRA48506.2021.9561151}
\newcommand\copyrightnotice{%
	\begin{tikzpicture}[remember picture,overlay]
	\node[anchor=south,yshift=10pt] at (current page.south) {\fbox{\parbox{\dimexpr\textwidth-\fboxsep-\fboxrule\relax}{\copyrighttext}}};
	\end{tikzpicture}%
}

\begin{document}

\maketitle
\thispagestyle{empty}
\pagestyle{empty}

\copyrightnotice

\vspace{-0.3cm}

\begin{abstract}

Solving robotic navigation tasks via reinforcement learning (RL) is challenging due to their sparse reward and long decision horizon nature. However, in many navigation tasks, high-level (HL) task representations, like a rough floor plan, are available. Previous work has demonstrated efficient learning by hierarchal approaches consisting of path planning in the HL representation and using sub-goals derived from the plan to guide the RL policy in the source task. However, these approaches usually neglect the complex dynamics and sub-optimal sub-goal-reaching capabilities of the robot during planning. This work overcomes these limitations by proposing a novel hierarchical framework that utilizes a trainable planning policy for the HL representation. Thereby robot capabilities and environment conditions can be learned utilizing collected rollout data. We specifically introduce a planning policy based on value iteration with a learned transition model (VI-RL). In simulated robotic navigation tasks, VI-RL results in consistent strong improvement over vanilla RL, is on par with vanilla hierarchal RL on single layouts but more broadly applicable to multiple layouts, and is on par with trainable HL path planning baselines except for a parking task with difficult non-holonomic dynamics where it shows marked improvements.
\end{abstract}

\section{Introduction}
\label{sec:intro}

In recent years, applying reinforcement learning (RL) to sequential decision making problems, has led to some remarkable successes like playing games with super-human performance \cite{mnih2015human, silver2018general} and mastering robotic manipulation \cite{andrychowicz2017hindsight} or locomotion \cite{schulman2015high} tasks. However, application of RL to robotic navigation tasks like avoiding obstacles and adapting the motion to environmental influences, in order to reach a goal, is still challenging. This is due to the long decision horizon and the naturally sparse reward that is only received for successful goal-reaching. Frequently used proxy rewards such as negative Euclidean distance to the goal can result in undesired side effects, like trapping the robot at obstacles. Therefore, several orthogonal concepts have been suggested to improve data-efficiency in sparse reward RL, including curriculum learning \cite{florensa2017reverse, klink2019selfpaced, wohlke2020performance}, advanced exploration \cite{osband2016deep, hong2018diversity}, or intrinsic motivation \cite{pathak2017curiosity, haber2018learning}. Our approach is related to hierarchical RL (HRL) \cite{bacon2017option, nachum2018data, vezhnevets2017feudal, levy2017learning} tackling long decision horizons by using a hierarchal policy.

\begin{figure}[htb]
	\centering
	\includegraphics[width=7cm]{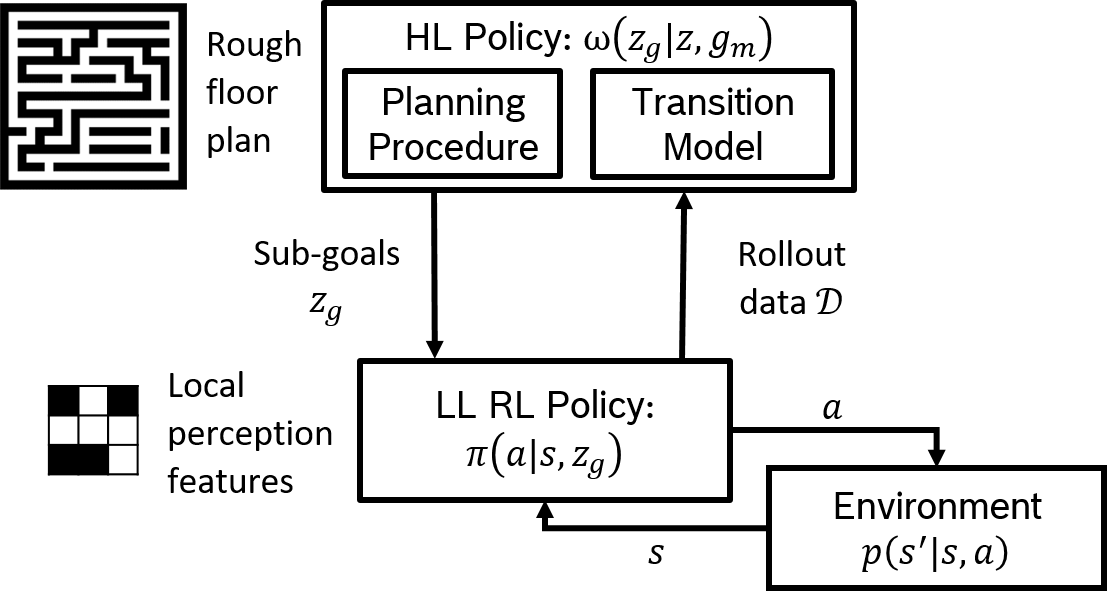}
	\caption{Our Hierarchical Planning and RL Framework}
	\label{fig:framework}
	\vspace{-0.3cm}
\end{figure}

In these ``vanilla" HRL approaches, abstraction is mainly achieved temporally by choosing different time scales for the hierarchy levels. Furthermore, the literature mainly focused on solving single instance tasks, like a specific environment layout \cite{nachum2018data}. In real world robotic navigation tasks, the environment layout or goal location might be different each time. At the same time, we often have additional information such as a rough floor plan with the goal location available. Using sampling-based planners \cite{faust2018prm, chiang2019rl} to obtain waypoints led to impressive real world results \cite{francis2020long} but a simulator that accurately represents the full navigation state space needs to be queried during planning.

While the task might be very complex in the original, continuous ``low-level" (LL) state and action space, including the full robot and environment dynamics, we observe that the task is much easier in an abstract ``high-level" (HL) floor plan representation where we can easily determine positional sub-goals. In these cases, previous work \cite{eppe2019semantics, yamamoto2018hierarchical} has demonstrated efficient learning by hierarchal approaches consisting of planning in the abstract HL representation and guiding the RL agent in the source task by sub-goals derived from the plan. However, they did not adapt the HL planning to the LL policy capabilities. In robotic navigation tasks, it is not realistic to assume that the robot can always reach a proposed sub-goal since the given HL representation might neglect important information about the robot or environmental dynamics as well as the current capabilities of the LL policy that affects the sub-goal-reaching.

Formalizing the assumptions about continuous state and action space robotic navigation tasks given a rough floor plan of the environment, we derive a novel hierarchical framework (see Fig.~\ref{fig:framework}) that overcomes these limitations by utilizing a trainable reactive policy for planning in the HL representation instead of purely conducting shortest path planning. This has the benefit that failures of the LL policy to reach the sub-goals are considered by adapting the HL planning utilizing gathered rollout data. Differentiable planning modules approximating value iteration (VI) \cite{tamar2016value, nardelli2018value} suggest simple integration with RL but have limitations with respect to stability with longer planning horizons \cite{tamar2016value} or properties of the assumed transition model \cite{nardelli2018value}. Therefore, we additionally present a HL planning policy based on exact VI featuring a learned transition model.

The key contributions of our work are:
\begin{itemize}
	\item Proposing a novel HRL framework derived from stated assumptions regarding domain knowledge combining planning in a HL state space with sub-goal guided RL in the original state space. This framework encompasses both the previous shortest path HL planning approaches as well as novel, trainable HL policies.
	\item Presenting a novel framework instance employing value iteration in the HL state space using a transition model learned from data gathered during LL RL (VI-RL).
	\item A detailed analysis in which scenarios and specific implementation learned HL policies provide benefits over pure shortest path plans.
\end{itemize}

We will work out these contributions by investigating the following research hypothesis:

\textit{H.1: Hierarchically combining planning in an abstract HL representation with goal-conditioned RL significantly increases the data-efficiency over flat ``vanilla" RL in continuous dynamics, sparse reward, long horizon tasks.}

\textit{H.2: Learning a HL transition model reflecting the RL policy's capabilities is necessary to handle environmental influences, or complex, non-holonomic dynamics.}

\textit{H.3: Transition models should consider both the HL state and action (chosen sub-goal); purely state-dependent models are insufficient to model non-holonomic constraints.} 


\section{Related Work}
\label{sec:related_work}

First ideas to generate sub-goals and hierarchically combine policies date back more than two decades \cite{schmidhuber1991learning, dayan1993feudal, sutton1999between}. Recent HRL approaches like Option Critic \cite{bacon2017option}, Feudal Networks (FuN) \cite{vezhnevets2017feudal}, or HiRO \cite{nachum2018data} make use of these ideas combining them with modern neural network policy architectures and RL algorithms: FuN and HiRO combine a HL policy that set sub-goals at a low frequency with a LL policy that receives intrinsic rewards for reaching these sub-goals. However, conventional HRL approaches do not consider available additional task knowledge.

Using sampling-based planners on a given map with reachability estimators learned from training experience to set sub-goals for local RL navigation policies were proposed in \cite{faust2018prm} (PRM) and \cite{chiang2019rl} (RRT).  However, an accurate simulator of the environment that allows for evaluating configurations in the original state space is necessary to execute the planning.

Several works combine planning on an abstract level with LL RL. They use different abstract HL representations like a symbolic representation \cite{yamamoto2018hierarchical, lyu2019sdrl}, an abstract representation based on a predicate logic \cite{eppe2019semantics}, or skills learned in an unsupervised fashion \cite{sharma2019dynamics}. The approaches presented in \cite{eppe2019semantics, yamamoto2018hierarchical} do not adapt their HL planning to the LL RL policy capabilities. In \cite{sharma2019dynamics}, the skills are not further updated in the target task and might therefore be bound to the environment and specific conditions under which they were discovered. While in \cite{lyu2019sdrl} the HL planning is adapted, the approach was only demonstrated for discrete state and action spaces but not for complex continuous ones relevant for robotic navigation.

Instead of pre-defining the HL representation, several works suggest ways to generate them from data: \cite{eysenbach2019search} built an abstract representation of the task by connecting collected experience in the replay buffer to a graph using a learned reachability estimator. A series of sub-goals is then planned by searching the obtained graph for the shortest path to the goal. This requires that the goal state can be connected to the obtained experience. In navigation tasks with large layouts where the goals state can be far away from experienced states or novel layouts this is not necessary feasible. Other approaches for learning plannable representations \cite{kurutach2018learning, corneil2018efficient, kipf2019contrastive, van2020plannable} struggle with similar challenges. Nevertheless, combining this orthogonal line of work with our hierarchical framework could be interesting future research.

Instead of learning a HL representation, various work in robotic navigation has focused on using an a-priori provided map. In \cite{zivkovic2006hierarchical} a base-level map is segmented to obtain a higher-level map representation for planning. Value iteration (VI) is carried out on different levels. In \cite{ma2020hierarchical} sub-environments are dynamically constructed as series of spatial-based state abstractions such that a tessellation of the environment results. Gupta et al. \cite{gupta2017cognitive} map first-person views to a top-down belief map of the world to apply a differentiable neural network planner on it. These works do not consider the continuous LL state space underlying robotic navigation problems and specifically for \cite{zivkovic2006hierarchical} do not learn the transition model for VI but assume an optimistic one.

It is, however, crucial to adapt the planning to the robot capabilities and the environmental dynamics, using rollout data, when combining sub-goal planning with RL. Here, differentiable planning modules \cite{tamar2016value, oh2017value, nardelli2018value, srinivas2018universal} address this issue. Requiring a discrete, ideally grid-like, environment representation, they enable learning value-function-based planning via backpropagation. The Value Iteration Network (VIN) \cite{tamar2016value} is a CNN architecture to approximate several recursions of VI planning. It has been generalized to arbitrary graphs \cite{niu2018generalized} and multiple levels of hierarchical planning \cite{schleich2019value}. The instability issues, especially with respect to long planning horizons, due to the recurrent structure are addressed in \cite{lee2018gated} and \cite{nardelli2018value}. MVProp \cite{nardelli2018value} mitigates unstable iteration by moving all trainable parameters into purely state-dependent propagation factors (normalized to $\lbrack 0, 1\rbrack$). This implicit assumption on the transition dynamics can harm the performance, for example in non-holonomic navigation tasks.

\section{Problem Statement}
\label{sec:problem_statement}

In this work, we consider MDPs $m$ sampled from a distribution of Markov Decision Processes (MDPs) $\mathcal{M}$. All MDPs $m$ share the same state and action space $\mathcal{S}$ and $\mathcal{A}$, the same goal-based reward function $r\left(s,g_m\right)$ with a goal $g_m$ sampled from a goal space $g_m \sim \mathcal{S}_{g,m}\subseteq \mathcal{S}$, and the same time horizon $T$ as well as discount factor $\gamma$. Each MDP $m$ has initial states $s_0$ sampled from an initial state space $s_0 \sim \mathcal{S}_{0,m}\subseteq \mathcal{S}$. Furthermore, each MDP has its own dynamics $\mathcal{P}_m$, but we assume that the dynamics are locally the same, i.e. the probability of a state transition is given by the same function of MDP instance specific features values $\mathcal{P}_m\left(s'\vert a,s\right)=f\left(\phi_{s,m},\phi_{s',m},\phi_{a,m}\right)$. 

The task is to find a goal-conditioned policy $\pi\left(a\vert s,g_m\right)$ that maximizes the expected returns under the distribution of MDPs, the goal and initial state distributions, and the dynamics:
\begin{align}\label{eq:objective}
\max_{\pi\left(a\vert s,g_m\right)} \mathbb{E}_{m \sim \mathcal{M}, g_m \sim \mathcal{S}_{g,m}, s_0 \sim \mathcal{S}_{0,m}, \mathcal{P}_m, \pi }\left[\sum^T_{t=0} \gamma^t r\left(s_t,g_m\right)\right]
\end{align}

We focus on sparse reward signals, which are common for robotic navigation problems. Therefore, the reward function is of the following form: $r\left(s, g_m\right) = \mathbb{I}_{d\left(s,g_m\right) \leq \epsilon} + c$ with $d\left(\cdot,\cdot\right)$ being a (weighted) distance function, $\epsilon$ the tolerance, and $c$ a constant to potentially offset the reward signal.

{\bf Assumptions on Domain Knowledge:} We will first explain at a robotic navigation task example the additional assumptions made to utilize a rough floor plan in our hierarchical framework. Afterwards, we will formally state them. As we will see later, these assumptions can often be fulfilled in practical applications and making use of them enables solving Eq.~\ref{eq:objective} more data-efficiently (see Sec.~\ref{sec:experimental_evaluation}).
\begin{figure}[htb]
\begin{center}
	\includegraphics[scale=0.8]{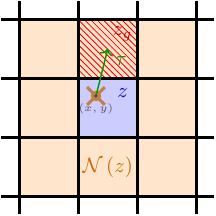}
	\caption{Exemplary High-Level State Space Illustration}
	\label{fig:ant_example}
\end{center}
\end{figure}

Imagine an ant-like robot navigating a continuous dynamics environment as depicted in Fig.\ref{fig:ant_example}. Given a rough floor plan, we can obtain an abstract, finite HL state space $\mathcal{Z}$ by partitioning the map into the shown grid of cells. Localizing the robot in $x,y$ coordinates, the corresponding grid cell (shown in blue) determines the current HL state $z$ (\textit{Asm.1}). In order to plan the next sub-goal, we need to have knowledge about local HL state neighborhood relations. In our example, all adjacent grid cells are the neighboring HL states $\mathcal{N}\left(z\right)$ (\textit{Asm.2}). Given a sub-goal $z_g$ in the HL state space (shaded in red), we need a way to condition our local LL sub-goal navigation policy acting in the original continuous state space on the HL sub-goal. Therefore a target vector $\tau$ (green) is calculated by taking the difference between the center of a grid cell and the robots' $x,y$ coordinates such that $s+\tau\left(s,z\right)$ is the LL sub-goal (\textit{Asm.3}). The HL state neighborhood $\mathcal{N}\left(z\right)$ must contain at least all directly reachable HL states. Else there is a risk that the only existing route to the goal is missed due to the state space abstraction (\textit{Asm.4}). To take into account and learn the robots capabilities for navigating various layouts, we need to be able to model HL state transitions using local features. In case of our grid representation, integer values indicating the presence or absence of obstacles or the terrain type for the (eight) neighboring grid cells $z' \in \mathcal{N}\left(z\right)$ can serve as such local map features $\psi_{z,m}$ (\textit{Asm.5}).

In summary, we formally assume:

\textit{Asm.1: There exists a finite HL state space $\mathcal{Z}$ with known $z=f_\mathcal{Z}\left(s\right)$.}

\textit{Asm.2: The neighbors $\mathcal{N}(z)$ of a HL state $z$ are known.}

\textit{Asm.3: There exists a function returning for a state $s$ and a HL state $z$ a target vector  $\tau\left(s,z\right)=f_\tau\left(s,z\right)$ such that $f_\mathcal{Z}\left(s+\tau\left(s,z\right)\right) = z$.}

\textit{Asm.4: $\mathcal{N}(z)\subset \mathcal{Z}$ contains at least all HL states directly reachable from $z$.}

\textit{Asm.5: Local features $\psi$ of any HL state are available. The probability of transitioning from $z$ to $z'\in \mathcal{N}\left(z\right)$ under any policy conditioned on a goal $z_g=f_\mathcal{Z}\left(g\right)$ can be approximated: $p_{\mathcal{P}_m,\pi\left(a\vert s,g\right)}\left(z'\vert z, z_g\right) \approx g\left(\psi_{z,m},\psi_{z',m},\psi_{z_g,m}\right)$.}

\section{Hierarchical Policy and Learning Framework}
\label{section:hierarchical_policy_and_learning_framework}

We introduce a novel policy and learning framework that efficiently solves Eq.~\ref{eq:objective} by utilizing the domain knowledge formalized in the assumptions stated above. It consists of:
\begin{itemize}
	 \item A \textit{goal-conditioned HL policy} $\omega\left(z_g\vert z,g_m\right)$ that selects a HL sub-goal $z_g \in \mathcal{N}\left(z\right)$ from the set of neighbors. This policy generally operates at a lower frequency than the LL state space transitions: A new sub-goal $z_g$ is selected if either the previous is reached or the number of time steps exceeds a threshold $T_{z_g} \ll T$.
	\item A \textit{sub-goal-conditioned LL policy} $\pi\left(a\vert s, f_\tau\left(s,z_g\right)+s\right)$ pursuing the sub-goals $\tau(s,z_g)+s$ set by the HL policy, instead of the MDP goal state $g_m$. In order to generalize across sub-goals as well as MDPs $m$, we model this policy by a function of the target vector and state features $\pi\left(a\vert s, f_\tau\left(s,z_g\right)+s\right)=h\left(a,f_\tau\left(s,z_g\right), \psi_{z,m}\right)$.
\end{itemize}

We assume only some domain knowledge, like a rough floor plan, to be given. The complex continuous LL dynamics of the robot interacting with its environment are unknown. 

In order to handle these unknowns, we introduce a hierarchical framework that allows training the HL sub-goal planning and the LL policy jointly using the same experience.

\subsection{Learning the Low-Level (LL) Policy}

During a rollout $\{\left(s_t,a_t,r_t, s_{t+1}\right)_{t=1,\ldots,T}\}$ in the environment, we track the corresponding HL states and actions $\{(z_t,z_{g,t},  z_{t+1})_{t=1,\ldots,T}\}$ as well as the time steps when a new HL sub-goal $z_{g, i}$ is selected $\{\left(t_i\right)_{i=1,\ldots,n}\}$. We train a LL sub-goal conditioned policy $\pi_\theta\left(a\vert s, f_\tau\left(s,z_g\right)+s\right)$ with parameters $\theta$ using the following intrinsic sub-goal-returns:
\begin{align}
\hat{R}_i=\sum^{t_{i+1}}_{t=t_i} \gamma^{t-t_i} \mathbb{I}_{f_\mathcal{Z}\left(s_{t+1}\right)=z_{g,t}},
\end{align}
An optimal LL policy with respect to these intrinsic rewards results in \emph{recursive optimality} \cite{ghavamzadeh2002hierarchically}. This means that the LL policy is only ``locally" optimal with respect to its sub-goal $z_g$. Even with an optimized HL policy the combined HL + LL policy can be sub-optimal with respect to Eq.~\ref{eq:objective} but will due to \textit{Asm.2} not miss an existing path to the goal. However, smaller ``sub-episodes" of maximum length $T_{z_g}$ improve the data-efficiency and enable generalization across layouts \cite{ghavamzadeh2002hierarchically}. While we employ TRPO \cite{schulman2015trust} for LL policy optimization, in our experiments, any other RL algorithm can be used as well.

\subsection{Learning the High-Level (HL) Policy}
\label{sec:instances}

{\bf Value iteration planning with explicit dynamics model (learning) [VI-RL]:} The discrete nature and small size of the HL state space $\mathcal{Z}$ and the HL action space $\mathcal{Z}_g$ combined with the known neighbor relations render value iteration (VI) tractable. To apply VI, we derive the HL reward function $r_\mathcal{Z}\left(z\right)=\mathbb{I}_{z=f_\mathcal{Z}\left(g_m\right)}+c$ from the given goal-based reward function of the MDP. Furthermore, we need a model of the HL state transitions $p_{\pi_\theta}\left(z'\vert z,z_g\right)$ that takes into account the dynamics induced by the sub-goal-conditioned LL policy $\pi_\theta\left(a\vert s, f_\tau\left(s,z_g\right)+s\right)$. We consider two different approaches for obtaining such a model:

\textit{Optimistic Model [VI-RL OM]:} Similar to \cite{eppe2019semantics, zivkovic2006hierarchical} this model optimistically assumes that a sub-goal $z_g$ always results in a transition to the corresponding HL neighbor: $p_{\pi_\theta}\left(z'\vert z,z_g\right)\approx \mathbb{I}_{z'=z_g}$. Using this model results in pure shortest path planning in grid-world domains.

\textit{Learned Model [VI-RL]:} Using the collected HL transition data $\{\left(z_{t_i},z_{g,{t_i}},  z_{t_{i+1}}\right)_{i=1,\ldots,n}\}$, we can learn a state feature-based model $p_{\pi_\theta}\left(z'\vert z,z_g\right)\approx \hat{g}_\lambda\left(\psi_{z,m},\psi_{z',m},\psi_{z_g,m}\right)$, with parameters $\lambda$, of the transitions. As we assumed these features to govern the HL dynamics, expressive models like deep neural networks should be able to closely approximate the dynamics. This has the benefit that the HL policy takes into account the LL policy capabilities. We train the model to maximize the sum of the log-likelihoods $\log \hat{g}_\lambda\left(\psi_{z,m},\psi_{z',m},\psi_{z_g,m}\right)$ regarding the observed data.

Using the reward function $r_\mathcal{Z}$ and the HL dynamics model $p_{\pi_\theta}$, we employ VI to obtain value functions $V\left(z\right),Q\left(z,z_g\right)$ that are used by the HL policy for sub-goal selection: $\omega\left(z_g\vert z,g_m\right)=\mathbb{I}_{z_g=\arg \max_{z' \in \mathcal{N}\left(z\right)} Q\left(z,z'\right)}$. During training, we employ an epsilon-greedy-variant of this policy for rollouts and use a replay-buffer for de-correlating training data, which we empirically found to result in more robust training. See Alg.~\ref{alg:vi-rl} for algorithmic implementation details of VI-RL.
\begin{algorithm}
	\DontPrintSemicolon
	\KwInput{Distribution of MDPs $\mathcal{M}$, domain knowledge ($\mathcal{Z}$, functions $f_\mathcal{Z}\left(s\right)$, $f_\tau\left(s, z\right)$, and $\mathcal{N}\left(z\right)$, and features $\psi_{z, m}$), HL transition model $\hat{g}_\lambda$, LL goal-conditioned policy $\pi_{\theta}$, HL and LL experience buffers $D_H$, $D_L$, LL horizon $T_{z_g}$.}
	
	\For{$j=1$ to $j_{\text{max}}$}
	{
		\While{$\vert D_L \vert < \textit{TRPO batch size}$}
		{
			\emph{\# Sample environment, start, and goal}\\
			$m \sim \mathcal{M}, s_{0,m} \sim \mathcal{U}\left(\mathcal{S}_{0,m}\right), g_m \sim \mathcal{U}\left(\mathcal{S}_{g,m}\right)$\\
			\emph{\# Conduct value iteration (VI)}\\
			$V\left(z\right), Q\left(z, z'\right) = \VI\left(\mathcal{Z}, \mathcal{N}\left(z\right), \hat{g}_\lambda, g_m\right)$\\
			\emph{\# Obtain HL policy}\\
			$\omega\left(z_g\vert z,g_m\right)=\mathbb{I}_{z_g=\arg \max_{z' \in \mathcal{N}\left(z\right)} Q\left(z,z'\right)}$\\
			\emph{\# Carry out RL episode}\\
			$t=0$, $i=0$ $t_0 = 0$, $z_{t_0} = f_\mathcal{Z}\left(s_{0,m}\right)$\\
			$z_{g,t_0} \sim \epsilongreedy\left(\omega\left(z_{t_0}\right)\right)$\\
			\While{$t < T$ and $d\left(s_t,g_m\right) > \epsilon$}
			{
				$a_t \sim \pi_\theta\left(s_t, f_\tau\left(s_t, z_{g, t_i}\right)\right)$\\
				$s_{t+1} \sim \mathcal{P}_m\left(s_t, a_t\right)$, $r_t =  \mathbb{I}_{f_\mathcal{Z}\left(s_{t+1}\right)=z_{g,t_i}}$\\
				$D_L \leftarrow \lbrace s_t, a_t, r_t, s_{t+1} \rbrace$\\
				\If{$\left(f_\mathcal{Z}\left(s_t\right)=z_{g, t_i}\right) \lor \left(t - t_i = T_{z_g}\right)$}
				{	
					\emph{\# Sub-goal reached}\\
					$t_{i+1} = t$, $z_{t_{i+1}} = f_\mathcal{Z}\left(s_{t+1}\right)$\\
					$D_H \leftarrow \lbrace z_{t_i}, z_{g, t_i}, z_{t_{i+1}} \rbrace$\\
					$z_{g, t_{i+1}} \sim \epsilongreedy\left(\omega\left(z_{t_{i+1}}\right)\right)$\\
					$i \leftarrow i+1$\\
				}
			}
			\emph{\# Update transition model using Adam \cite{kingma2014adam}}\\
			$\lambda_{j+1} \leftarrow \Adam\left(\lambda_j, D_H\right)$\\
		}
		\emph{\# Update RL policy}\\
		$\theta_{j+1} \leftarrow \TRPO\left(\theta_j, D_L\right)$, $D_L \leftarrow \emptyset$
	}
	\caption{VI-RL with LL TRPO Policy}
	\label{alg:vi-rl}
\end{algorithm}

{\bf Differentiable planning module with implicitly learned dynamics model (MVPROP-RL):} In RL tasks with a discrete, grid-like state space representation, a differentiable recursive architecture like MVProp \cite{nardelli2018value} can be used to approximate the value function. In our setting, we can therefore employ MVProp to approximate the HL value function $V_\pi\left(z\right)$ with a model $T\,\hat{V}^K_\lambda\left(z\right)$ with $T$ being the time horizon.

$\hat{V}^K_\lambda\left(z\right)$ is the result of $K$ recursions where the value of neighboring states is propagated to state $z$ based on a state-dependent propagation factor $p_\lambda (z)$ representing simplified transition dynamics. We train the model $T\,\hat{V}^K_\lambda\left(z\right)$ with collected data $\{\left(z_{t_i},z_{g,{t_i}},  z_{t_{i+1}}\right)_{i=1,\ldots,n}\}$ using similar to \cite{nardelli2018value} a TD-$0$ loss (but omitting the importance weighting, which we empirically found unnecessary): $l\left(z,z',\lambda\right)= \left(\left(r\left(z\right) + \left(1 - \mathbb{I}_{f_\mathcal{Z}\left(g_m\right)=z}\right) \gamma T\,\hat{V}^K_{\lambda_\text{target}}\left(z'\right)\right) - T\,\hat{V}^K_\lambda\left(z\right)\right)^2$. 

We obtain the HL policy as $\omega\left(z_g\vert z,g_m\right)=\mathbb{I}_{z_g=\arg \max_{z' \in \mathcal{N}\left(z\right)}\hat{V}^K_\lambda\left(z'\right)}$. Similar to VI-RL we employ an epsilon-greedy exploration and a replay-buffer.


\section{Experimental Evaluation}
\label{sec:experimental_evaluation}

So far, we formalized assumptions for robotic navigation tasks given a rough floor plan and formulated a framework that aims at accelerating learning by utilizing this domain knowledge. In the following, we empirically investigate the research hypotheses stated in Sec.~\ref{sec:intro}.\footnote{Code is planned to be made available at https://github.com/boschresearch/Hierarchies-of-Planning-and-Reinforcement-Learning-for-Robot-Navigation is planned}

\subsection{Environments}
\label{sec:environments}

\begin{figure}[tb]
	\vspace{0.2cm}
	\centering
	\begin{subfigure}{.24\textwidth}
		\centering
		\includegraphics[width=2.5cm]{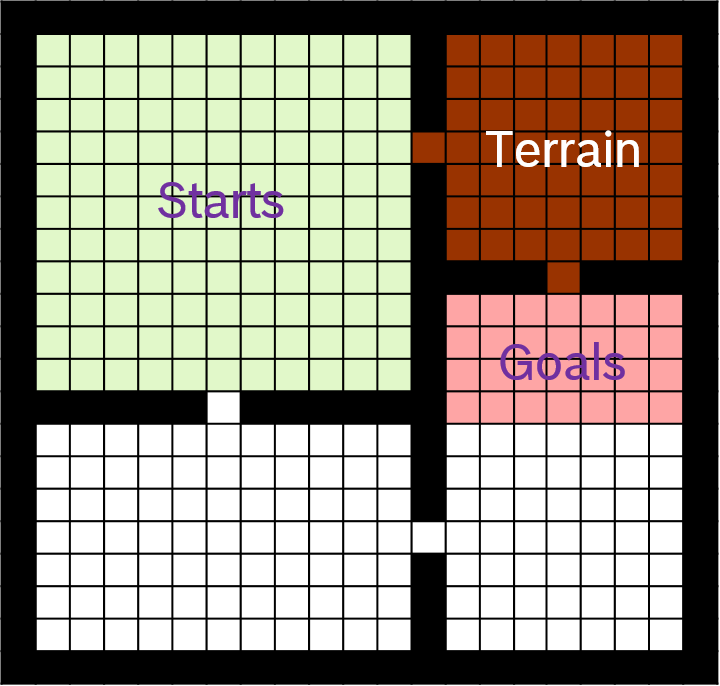}
		\caption{Four Rooms\\ $T=100$, $T_{z_g}=2$}
		\label{fig:4r_env}
	\end{subfigure}%
	\begin{subfigure}{.24\textwidth}
		\centering
		\includegraphics[width=2.5cm]{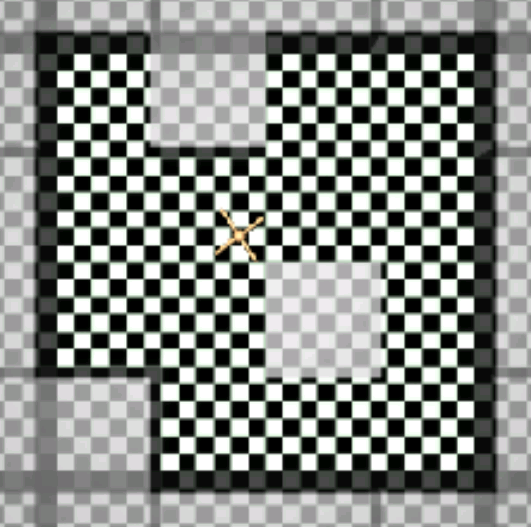}
		\caption{MuJoCo Ant Mazes\\ $T=2000$, $T_{z_g}=20$}
		\label{fig:ant_env}
	\end{subfigure}%

	\begin{subfigure}{.24\textwidth}
		\centering
		\includegraphics[width=2.5cm]{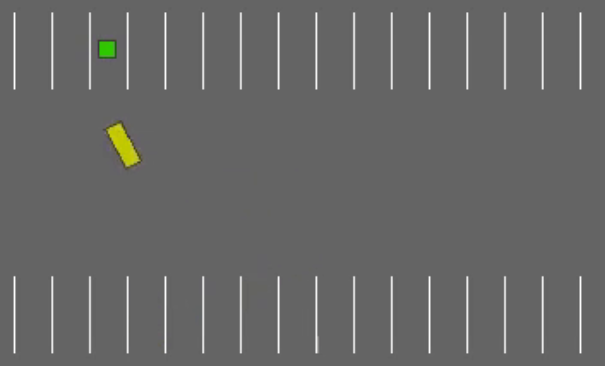}
		\caption{Vehicle Parking\\ $T=100$, $T_{z_g}=2$}
		\label{fig:parking_env}
	\end{subfigure}
	\caption{Simulation Environments Overview}
	\label{fig:simulation_envs}
\end{figure}

We consider three simulated continuous dynamics robotic navigation domains, depicted in Fig.~\ref{fig:simulation_envs}, which demonstrates the broad feasibility of the assumptions (Sec.~\ref{sec:problem_statement}). The use of simulation enables us to perform the number of training rollouts and independent trainings (seeds) required by RL.

{\bf Four Rooms:} A continuous dynamics ($x,y$-position and -velocity) point mass robot needs to navigate through the environment in Fig.~\ref{fig:4r_env}. Starting positions are sampled in the top left, whereas goal locations are sampled in the upper part of the bottom right room. The continuous $x,y$-accelerations are the LL control inputs. The discrete HL state representation results from tiling the layout into the shown $21 \times 21$ tiles. The top-right room features terrain that uniformly slows down the robot motion. For the HL transition model of VI-RL, we train an MLP architecture. It receives HL state features, the terrain information (free, wall, terrain) for the eight neighbors, and an integer indicating the relative neighbor chosen as sub-goal, as inputs. A vector of transition probabilities for the eight neighbors is the output.

\begin{figure*}[tb]
	\vspace{0.2cm}
	\centering
	\begin{subfigure}{.31\textwidth}
		\centering
		\includegraphics[width=6cm]{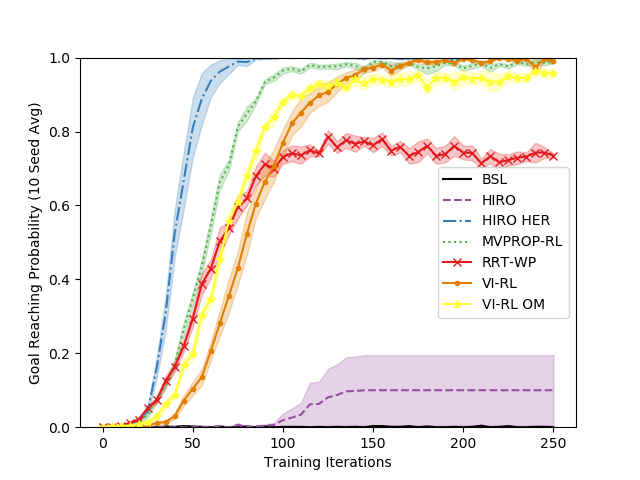}
		\caption{Four Rooms (No Terrain)}
		\label{fig:results_4r_nt}
	\end{subfigure}
	\begin{subfigure}{.31\textwidth}
		\centering
		\includegraphics[width=6cm]{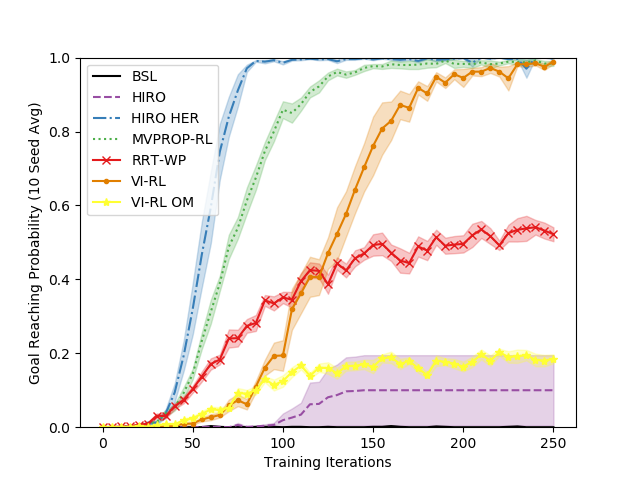}
		\caption{Four Rooms}
		\label{fig:results_4r_t}
	\end{subfigure}%
	\begin{subfigure}{.31\textwidth}
		\centering
		\includegraphics[width=6cm]{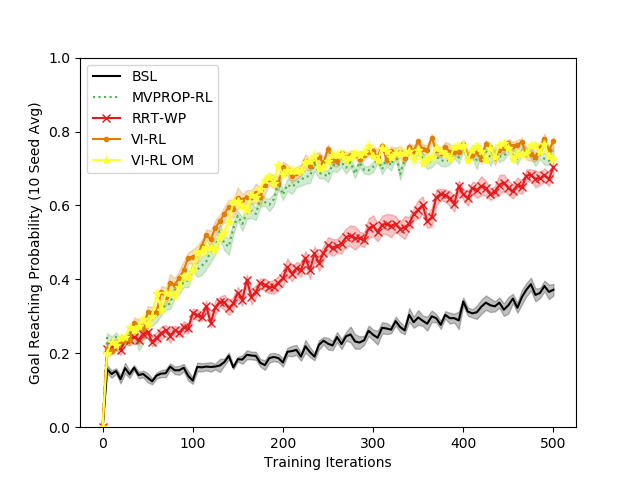}
		\caption{MuJoCo Ant Mazes}
		\label{fig:rm_ant}
	\end{subfigure}
	\caption{Maze Navigation; Line: Mean across 10 Seeds; Shaded Area: Standard Error}
	\label{fig:rexperiments}
\end{figure*}

{\bf MuJoCo Ant Mazes:} A MuJoCo \cite{todorov2012mujoco} ant robot with complex continuous dynamics needs to navigate one of 25 randomly sampled $6 \times 6$ block environments with different obstacle structure. We obtain the HL state space by sub-dividing each block into four tiles ($24 \times 24$ overall). The resolution is a hyper-parameter to be reasonably chosen considering robot dimensions and sub-goal horizon $T_{z_g}$.

{\bf Vehicle Parking \cite{highway-env}:} A non-holonomic bicycle dynamics vehicle needs to park in a randomly selected slot, starting in the middle of the parking lot. Unlike the other domains, the robot orientation $\vartheta$ is part of the goal specification and an integral part of the state and action space. We obtain a 3D HL state space by tiling the $x,y$-position into $24 \times 12$ tiles and the orientation into eight $\SI{45}{\degree}$ segments. With no obstacles, any HL position has identical (terrain) features. Therefore, we remove these from the input of the HL transition model of VI-RL. Instead, we augment the in- and output to account for the $\vartheta$ segments. For MVPROP(-RL), we use a 3D-CNN. The robot receives a sparse reward for successfully parking.

\subsection{Baselines}
\label{sec:baselines}

{\bf Vanilla RL (BSL):} A flat, non-hierarchical architecture, using the goal-conditioned RL policy with a target vector to the final MDP goal $g_m$ instead of a sub-goal target vector.

{\bf RRT-Waypoints (RRT-WP):} Inspired by \cite{chiang2019rl}, an RRT planner generates a path of waypoints from start to goal in 2D $x,y$-space using the map. The LL goal-conditioned RL policy is trained to reach the waypoint selected as sub-goal.

{\bf HiRO} \cite{nachum2018data} is a state-of-the-art HRL approach. It uses a conventional MLP HL policy, trained via off-policy RL (TD3 \cite{fujimoto2018addressing}) and operating in the continuous state space like the LL policy. The LL policy is trained to follow target vectors generated (at the same frequency as in VI-/MVPROP-RL) by the HL. While in \cite{nachum2018data} both policy levels use distance-based reward shaping, we provide the HL policy with the environmental sparse rewards to ensure comparability with the other approaches and avoid trapping the robot at obstacles.

Aiming for a fair comparison, we use the same number of hierarchy levels (except BSL) and the same goal-conditioned LL TRPO \cite{schulman2015trust} policy architecture for all approaches.

\subsection{Simple Point-Mass Navigation}
\label{sec:simple_point_mass_navigation}

In order to investigate under which circumstances learning HL transition dynamics implicitly (MVPROP-RL) or explicitly (VI-RL) is necessary (see research hypothesis \textit{H.2}), we conduct experiments in a simplistic environment with a focus on the relative effects of environmental conditions instead of absolute performance. We compare the goal-reaching probability across training for \textit{Four Rooms} domain with (Fig.~\ref{fig:results_4r_t}) and without (Fig.~\ref{fig:results_4r_nt}) the hard-to-traverse terrain in the top-right room. The distance-wise shortest paths from the starts to the goals traverse the top-right room. With the terrain, however, the optimal policy takes the ``longer" routes through the bottom-left room. Looking at the results, vanilla RL (BSL) is not able to reach the goals (\textit{H.1}). VI-RL and MVPROP-RL, like HIRO HER, achieve $\SI{100}{\percent}$ goal-reaching. Using the capacious transition model necessary for the more difficult realistic tasks (e.g. Sec.~\ref{sec:robotic_maze_navigation}), VI-RL learns a bit slower. Comparing VI-RL with the learned transition model to VI-RL OM  with the optimistic model, it turns out that the performance is similar without the terrain, whereas the goal-reaching of VI-RL OM drops to less than $\SI{20}{\percent}$ with the terrain, because the episodes `time out' while crossing the terrain. We witness a less severe performance drop of RRT-WP. Due to the sampling-based planning, by chance, a path through either the top-right or bottom-left room is found, resulting in $\SI{50}{\percent}$ goal-reaching. In conclusion, these evaluations in the simplistic environment show that learning the HL transition dynamics is crucial for reacting to environmental conditions, like terrain-type, which is necessary in realistic set-ups (\textit{H.2}).

\subsection{Robotic Maze Navigation}
\label{sec:robotic_maze_navigation}

In real application, a robot needs to be able to navigate a broad range of environment layouts. Therefore, we evaluate the algorithms on the \textit{MuJoCo Ant Mazes} environment, featuring 25 maze layouts randomly picked for every rollout. We excluded HiRO as it was not designed to solve multiple task instances\footnote{Additional experiments showed that replacing the HL MLP policy with a CNN does not successfully enable HiRO to handle multiple layouts.}. The results are depicted in Fig.~\ref{fig:rm_ant}. VI-RL and MVPROP-RL perform similar, clearly outperforming the vanilla RL baseline BSL (see research hypothesis \textit{H.1}). Surprisingly, using the optimistic model (OM) does not decrease the VI-RL performance, despite the very complex ant dynamics. It is apparently similarly difficult to move the ant in any specific direction throughout the training. Hence, purely planning the shortest path is sufficient in this static setting (\textit{H.2}). In this light, the slower learning of RRT-WP is explained by the RRT only approximating the shortest path.

\subsection{Non-Holonomic Vehicle Parking}

\begin{figure}[tb]
	\centering
	\includegraphics[width=6cm]{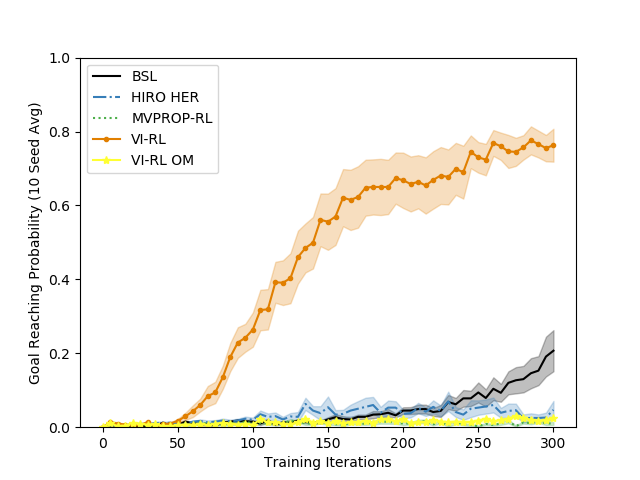}
	\caption{Vehicle Parking; Line: Mean across 10 Seeds; Shaded Area: Standard Error}
	\label{fig:parking}
\end{figure}

As previously stated, learning \textit{Vehicle Parking} adds the difficulty of non-holonomic dynamics to the learning task, involving a tight coupling of location and orientation. These non-holonomic constraints must be taken into account during HL planning, in order to master the task (see research hypothesis \textit{H.3}). The results are depicted in Fig.~\ref{fig:parking}. Apart from VI-RL, basically, all other approaches are not able to significantly increase the goal-reaching probability in reasonable time. VI-RL quickly picks up the pace and reaches over $\SI{80}{\percent}$ goal-reaching probability. This task provides another yet very different example to the \textit{Four Rooms} with terrain domain where learning the transition dynamics is crucial: The purely shortest path planning of VI-RL OM does not consider the non-holonomic constraints and therefore does not provide effective guidance (\textit{H.2}). Furthermore, the experiment shows the limitations of the assumptions underlying the transition model of MVProp \cite{nardelli2018value}: Only learning state-dependent propagation factors $p$ that do not take into account to which specific neighboring HL state the robot shall transition is not sufficient. Without any obstacles in this environment, in principal, all HL states are traversable, however, specific transitions like going `north' while the robot orientation is `west' may be difficult to execute. In contrast, the learned HL transition model in VI-RL explicitly takes these non-holonomic constraints into account and hence is able to guide the LL policy to reach the goal (\textit{H.3}).

\section{Conclusion}
\label{sec:conclusion}

In this work, we formalized the domain knowledge typically available in robotic navigation tasks, like a rough floor plan of the environment, into assumptions. Based on these, we derived a novel framework combining planning in a high-level (HL) state space with sub-goal guided RL in the original continuous state space. Furthermore, we presented a novel, specific instance of our framework building on value iteration planning in the HL state space using a transition model learned from data gathered during RL (VI-RL). 

In several difficult sparse reward robotic navigation tasks, we examined several research hypotheses with the following outcomes: (\textit{H.1}) Hierarchically combining planning and RL enables handling random task instances while achieving significant data-efficiency gains over non-hierarchical RL. (\textit{H.2}) Comparing different HL planning approaches, we found that in static 2D navigation environments learning HL transition models is, surprisingly, often not necessary and pure shortest path planning performs equally well. However, learning the dynamics becomes crucial when the navigation needs to take into account environmental influences like the terrain type. (\textit{H.3}) In the practically relevant non-holonomic vehicle parking task featuring a more complex HL state space, we uncovered limitations of the optimistic model as well as the purely state-dependent transition model of MVProp planning. Our novel algorithm VI-RL overcomes these limitations by employing a transition model accounting for the selected HL sub-goal state and significantly outperforms all baselines. 

In future work, an interesting direction of research could be to also learn the transformation of the original state space into the HL state space instead of designing it based on the domain knowledge. This way, the presented approaches could be even more broadly applicable.





\end{document}